\documentclass[10pt,letterpaper]{article}
\usepackage[top=0.85in,left=2.75in,footskip=0.75in,marginparwidth=2in]{geometry}

\usepackage[utf8]{inputenc}

\usepackage{cite}

\usepackage{nameref,hyperref}

\usepackage[right]{lineno}

\usepackage{microtype}
\DisableLigatures[f]{encoding = *, family = * }

\raggedright
\usepackage{changepage}

\usepackage[aboveskip=1pt,labelfont=bf,labelsep=period,singlelinecheck=off]{caption}

\makeatletter
\renewcommand{\@biblabel}[1]{\quad#1.}
\makeatother

\usepackage{lastpage,fancyhdr,graphicx}
\usepackage{epstopdf}
\fancyheadoffset[L]{2.25in}
\fancyfootoffset[L]{2.25in}

\usepackage{color}

\definecolor{gray}{gray}{.25}

\usepackage{graphicx}

\usepackage{sidecap}

\usepackage{wrapfig}
\usepackage[pscoord]{eso-pic}
\usepackage[fulladjust]{marginnote}
\reversemarginpar

\usepackage{enumitem,pdflscape}
\definecolor{caption_main}{gray}{0.25}
\definecolor{caption_sub}{gray}{0.25}
\definecolor{purple}{RGB}{128, 0, 128}
\usepackage{wrapfig}
\newcommand{\etal}{\textit{et al}. }
\newcommand{\ie}{\textit{i}.\textit{e}. }
\newcommand{\eg}{\textit{e}.\textit{g}. }
\newcommand{\citep}{\cite}
\renewcommand{\textdegree}{$^{\circ}$}
\begin{document}
\vspace*{0.35in}
\begin{flushleft}
{\Large
\textbf\newline{Multidimensional ground reaction forces and moments from wearable sensor accelerations via deep learning}
}
\newline
\\
William~R.~Johnson\textsuperscript{1,*},
Ajmal~Mian\textsuperscript{2},
Mark~A.~Robinson\textsuperscript{3},
Jasper~Verheul\textsuperscript{3},
David~G.~Lloyd\textsuperscript{4},
Jacqueline~A.~Alderson\textsuperscript{1,5}
\\
\bigskip
\bf{1} {School of Human Sciences (Exercise and Sport Science), The University of Western Australia, Perth, Australia.} 
\\
\bf{2} {Department of Computer Science and Software Engineering, The University of Western Australia, Perth, Australia.} 
\\
\bf{3} {Research Institute for Sport and Exercise Sciences (RISES), Liverpool John Moores University, Liverpool, England.}
\\
\bf{4} {Menzies Health Institute Queensland, and the School of Allied Health Sciences, Griffith University, Gold Coast, Australia.}
\\
\bf{5} {Sports Performance Research Institute New Zealand (SPRINZ), Auckland University of Technology, Auckland, New Zealand.}
\\
\bigskip
* {\href{mailto:bill@johnsonwr.com}{\color{blue}bill@johnsonwr.com}}
\vskip 6pt
\href{https://doi.org/10.1109/TBME.2020.3006158}{\texttt{\color{blue}https://doi.org/10.1109/TBME.2020.3006158}}
\end{flushleft}
\section*{Abstract}
\textit{Objective:} Monitoring athlete internal workload exposure, including prevention of catastrophic non-contact knee injuries, relies on the existence of a custom early-warning detection system. This system must be able to estimate accurate, reliable, and valid musculoskeletal joint loads, for sporting maneuvers in near real-time and during match play. However, current methods are constrained to laboratory instrumentation, are labor and cost intensive, and require highly trained specialist knowledge, thereby limiting their ecological validity and wider deployment. An informative next step towards this goal would be a new method to obtain ground kinetics in the field.
\textit{Methods:} Here we show that kinematic data obtained from wearable sensor accelerometers, in lieu of embedded force platforms, can leverage recent supervised learning techniques to predict near real-time multidimensional ground reaction forces and moments (GRF/M). Competing convolutional neural network (CNN) deep learning models were trained using laboratory-derived stance phase GRF/M data and simulated sensor accelerations for running and sidestepping maneuvers derived from nearly half a million legacy motion trials. Then, predictions were made from each model driven by five sensor accelerations recorded during independent inter-laboratory data capture sessions. 
\textit{Results:} The proposed deep learning workbench achieved correlations to ground truth, by maximum discrete GRF component, of vertical $F_z$ 0.97, anterior $F_y$ 0.96 (both running), and lateral $F_x$ 0.87 (sidestepping), with the strongest mean recorded across GRF components 0.89, and for GRM 0.65 (both sidestepping). 
\textit{Conclusion:} These best-case correlations indicate the plausibility of the approach although the range of results was disappointing. The goal to accurately estimate near real-time on-field GRF/M will be improved by the lessons learned in this study.
\vspace*{\fill}
\marginpar{Supplementary material available (\href{http://digitalathlete.org}{\texttt{\color{blue}digitalathlete.org}})}
\marginpar{\copyright2020 IEEE}
\newpage
\noindent
\textit{Significance:} Coaching, medical, and allied health staff could ultimately use this technology to monitor a range of joint loading indicators during game play, with the aim to minimize the occurrence of non-contact injuries in elite and community-level sports.\par
\bigskip
\noindent\textbf{Keywords} Biomechanics $\cdot$ Wearable sensors $\cdot$ Simulated~accelerations~$\cdot$ Workload~exposure~$\cdot$~Sports~analytics~$\cdot$~Deep~learning\par
\section{Introduction}
One of the perpetual problems facing sports biomechanists is the difficulty translating the accuracy and multidimensional fidelity of laboratory-based measurements and downstream analysis into the sporting arena \citep{5RN387chiari,5RN409elliott}. 
In pursuit of the monitoring of the multiple contributors to player welfare, of acute and chronic injury risk plus external and internal workload exposure \citep{5RN751gabbett2018,5RN753vanrenterghem2017}, coaches today are forced to make local interpretations of surrogate measures \citep{5RN734rossi,5RN729bradley,5RN718boudreaux}. 
Traditional outputs of biomechanical analyses, ground reaction forces and moments from embedded force plates and for example knee joint moments (KJM) from calculations of inverse dynamics, which could be considered candidate variables of interest to the monitoring ensemble, have so far been captive to the laboratory \citep{5RN751gabbett2018,5RN753vanrenterghem2017,5RN760matijevich,5RN769coutts2007,5RN777calvert}. 
Using catastrophic non-contact knee injuries as an example, there is a gap between the understanding of the mechanisms of anterior cruciate ligament injury, and the ability to monitor the collection of associated risk parameters during a game \citep{5RN750chinnasee,5RN703besier2001,5RN499dempsey2007,5RN747johnson2018arXiv}.\par
The traditional approach to biomechanical analysis begins with laboratory retro-reflective optical motion capture recorded in synchronization with analog force plate output \citep{5RN387chiari,5RN399lloyd2000}. The University of Western Australia holds a legacy archive of movement data, and this was considered an advantage and enabler for the current data science investigation. The major advantage of inertial measurement units (IMU) over optical motion capture is the relative ease of on-field application away from the laboratory, however, there are several limitations to the currently accepted linear processing of their telemetry output. An IMU typically contains three discrete devices: an accelerometer (linear acceleration); gyroscope (angular velocity); and magnetometer (to derive orientation) \citep{5RN708camomilla}. These IMU sensors are often used alongside global positioning system (GPS) trackers in a combined unit which allows positional information (facilitating game strategy and tactical analysis) to be included in workload exposure estimations \citep{5RN776hennessy,5RN743buchheit,5RN751gabbett2018,5RN753vanrenterghem2017}. In processing IMU outputs, linear statistics tend to be based on gross assumptions, which for example can mistake overfitting for personalization \citep{5RN719callaghan,5RN604wundersitz,5RN735gabbett,5RN734rossi,5RN729bradley}. Scientific investigation to employ IMU for movement classification and load estimation has so far shown more success with basic movements and/or unidimensional GRF analysis \citep{5RN722pham,5RN785verheul,5RN732clermont,5RN742thiel}. The IMU hardware also has inherent physical characteristics and design features which need to be carefully controlled. The three sensors have relative or independent coordinate systems, and vendors use proprietary algorithms based on Kalman filters \citep{5RN504karatsidis,5RN708camomilla,5RN740luinge} and custom orientation calibration \citep{5RN738lebel,5RN741picerno,5RN745lipton} to determine the device position with respect to the laboratory global origin. Both the accelerometer and gyroscope are susceptible to linear (or quadratic) drift depending on the application of integration calculations \citep{5RN708camomilla}. The magnetometer is affected particularly by the proximity of ferromagnetic materials which can be a problem with laboratory and field equipment \citep{5RN708camomilla,5RN736ancillao}. One common error is the misinterpretation of IMU results during treadmill activities where anteroposterior acceleration is naturally minimized \citep{5RN725wouda,5RN730koska,5RN737ngoh,5RN732clermont}.  Wearable devices are also prone to task-dependent fixation and skin artefacts, in other words powerful movement types necessitate a more stable attachment to the body, for example throwing or explosive change of direction activities, or any movement where the IMU is at the distal end of the moment arm \citep{5RN708camomilla,5RN504karatsidis}. All these issues are compounded when multiple devices are deployed per participant, each of which must be synchronized, and where bandwidth to a Bluetooth or Wi-Fi bridge is shared.  In a team situation, one of the most challenging problems is the logistics of managing the consistency of device hardware and software versions \citep{5RN726nicolella,5RN743buchheit}.
In short, IMU devices are often preferred over optical motion capture for ease of setup and preservation of ecological validity, however, their use comes with a set of constraints and limitations, some of which have remained difficult to solve \citep{5RN387chiari,5RN409elliott,5RN753vanrenterghem2017}.\par
An emerging alternative method of processing IMU data output is deep learning (or deep neural network, DNN), which is a type of an artificial intelligence system based on a learning model rather than a task-specific algorithm \citep{5RN717lecun}. The successful deployment of DNN machine learning for practical biomechanical applications benefits from a multidisciplinary sport science and computer science approach and early researchers have applied this technology with IMU to classify gait, predict vertical GRF ($F_z$); or segment orientation \citep{5RN724jacobs,5RN744hu,5RN736ancillao,5RN737ngoh,5RN727zimmermann,5RN725wouda}. Recent CNN models, \eg AlexNet and ResNet, are highly successful at classifying image contents \citep{5RN453krizhevsky2012,5RN752he2015}, and it is possible using fine-tuning (transfer learning) to leverage these existing CNNs for related applications and from fewer training samples (\ie thousands instead of millions) with concomitant reductions in CPU and GPU processing cost.\par
\begin{figure*}[t]
\begin{adjustwidth}{-2in}{0in}
    \begin{center}
    \includegraphics[width=0.9\linewidth]{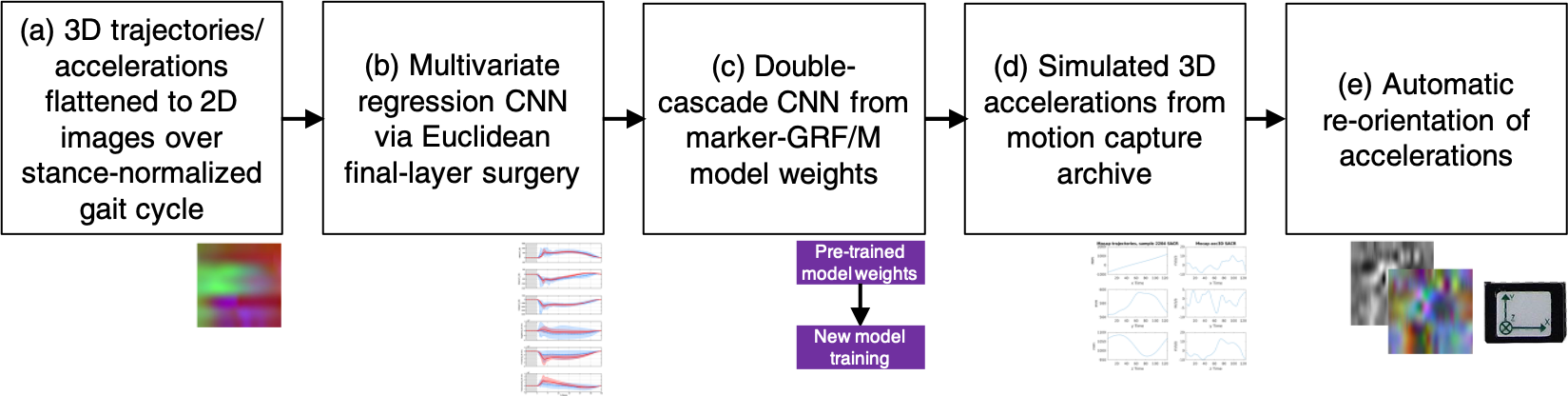}
    \end{center}
    \captionsetup{labelformat=empty}
    \caption{}
    \label{ch5_figA}
\end{adjustwidth}
\end{figure*}
\begin{figure*}[b]
\begin{adjustwidth}{-2in}{0in}
    \begin{center}
    \includegraphics[width=0.9\linewidth]{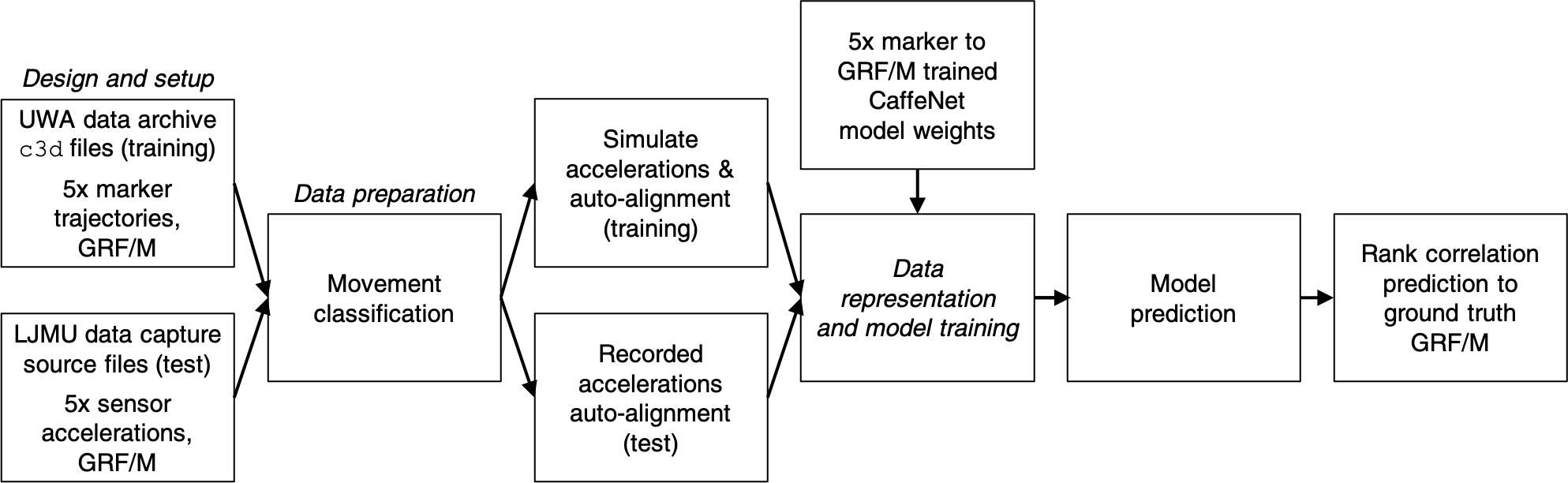}
    \end{center}
    \captionsetup{labelformat=empty}
    \caption{}
    \label{ch5_figB}
\end{adjustwidth}
\end{figure*}

\marginpar{
    \begin{flushleft}
    \vspace{-7.8cm}
    {\color{caption_main}\textbf{Figure \ref{ch5_figA}. Deep learning workbench for biomechanics.~{\color{caption_sub}The sequence of data science techniques used by the study. The practical application of these steps, and ultimate prediction of GRF/M waveforms, is described in the \textit{Data representation \& model training} subsection of the \textit{Methods}.}}}
    \end{flushleft}
}
\marginpar{
    \begin{flushleft}
    \vspace{1.5cm}
    {\color{caption_main}\textbf{Figure \ref{ch5_figB}. Study overall \mbox{design}.}}
    \end{flushleft}
}
A major step towards model deployment and acceptance in the field is proving its accuracy and validity in sub-optimal or adversarial conditions. Previous work has tested CNN models using a conventional 80:20 split of homogeneous archive movement data to predict three dimensional (3D) GRF/M and KJM from marker trajectories. This was achieved by building a ``deep learning workbench'' which (a) flattened 3D marker trajectories to 2D images in order to allow fine-tuning of image classification deep models; (b) transplanted Euclidean loss into the final CNN layer to facilitate multivariate regression; and (c) realized improvements in downstream KJM model accuracy by leveraging earlier GRF/M success \citep{5RN558johnson2018tbme,5RN747johnson2018arXiv} (Figure~\ref{ch5_figA}). The current investigation began by investigating model performance using a training-set of simulated accelerations, against a test-set of recorded sensor accelerations, both with corresponding GRF/M.  This required the workbench to be extended to (d) synthesize accelerations from marker trajectories, and (e) to automatically re-orient independent acceleration coordinate systems so that they are aligned with the global coordinates.\par
\begin{wrapfigure}{l}{0.55\textwidth}
    \fontsize{8.0}{8.0}\selectfont
    \renewcommand{\arraystretch}{1.4}
    \includegraphics[width=1.0\linewidth]{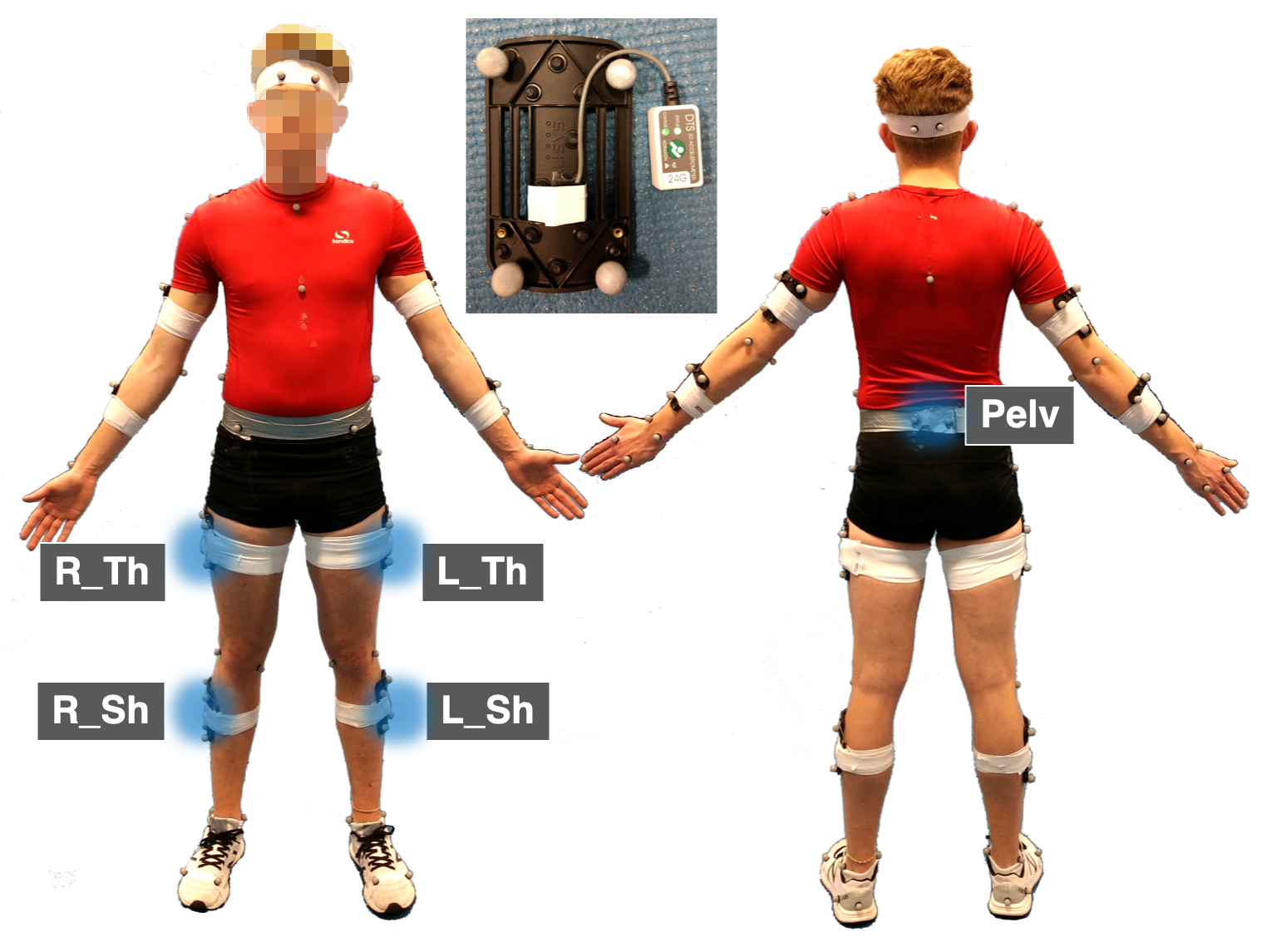}
    \captionsetup{labelformat=empty}
    \caption{}
    \label{ch5_figC}
\end{wrapfigure}

\marginpar{
    \vspace{0.5cm}
    {\color{caption_main}\textbf{Figure \ref{ch5_figC}. Location of five sensor accelerometers.}~{\color{caption_sub}Each sensor is shown artificially colored and labeled (LJMU naming convention). Inset, for the thigh and shank locations, the accelerometer was attached to a rigid plate.}}
}    
The contribution of this study is to investigate the resilience of the workbench when faced with a test-set of sensor accelerations recorded independently of the primary researcher (and inter-laboratory), thus providing a real-world scenario and reducing the possibility of home-game advantage or bias. Because the telemetry was provided by another laboratory, calibration parameters such as coordinate system, direction of travel, number, type and location of sensor accelerometers, even the make and model of the force plate systems, required subsequent data preparation and representation to be more generalized. Prediction analysis was carried out using the Caffe deep learning framework \citep{5RN579jia2014} on two different CNN models, CaffeNet (a derivative of AlexNet) and ResNet, both via double-cascade learning, using weights from earlier marker trajectories to GRF/M models, themselves fine-tuned from ImageNet source big data \citep{5RN453krizhevsky2012,5RN752he2015}. The CNN models were trained using accelerations simulated from an archive of marker trajectory data captured at The University of Western Australia (UWA, Perth, Western Australia), and tested with sensor accelerations recorded at Liverpool John Moores University (LJMU, Liverpool, UK). The accuracy and validity of the approach was tested by reporting correlations between CNN predicted and ground truth GRF/M over 100~\% of time-normalized stance for two sports-related movement patterns, running and sidestepping. The hypothesis was that CNN models can establish the location of sensor accelerometers via the signature pattern of 3D accelerations, and that this would be demonstrated by mean GRF and GRM correlations $>~0.80$ across all movement types and stance limb combinations. It was anticipated that the results of this study would add to the understanding of the performance of CNN models driven by 3D accelerations, and contribute to future practitioners' placement of sensor accelerometers for optimum results.\par
\section{Methods}
\subsection{Design \& setup}
The overall design of the study is shown in Figure~\ref{ch5_figB}. For the current investigation, the training and test data were from different sources. A UWA archive of marker trajectories and GRF/M data from a 17--year period from 2001--2017 was used to train the CNN models.  Gathered from multiple biomechanics laboratories, the training data files selected from the total of 458,372 shared common optical motion capture setup (12--20 Vicon camera models MCam2, MX13 and T40S; Oxford Metrics, Oxford, UK), analog force plate configuration (Advanced Mechanical Technology Inc, Watertown, MA), data capture software (Vicon Workstation v4.6 to Nexus v2.5), and young adult athletic participant cohort (male $59.9~\%$, female $40.1~\%$, height $1.770~\pm~0.101~m$, and mass $74.9~\pm~34.1~kg$). The UWA optical marker set has varied over this period from 24--67 passive retro-reflective markers.  However, for this investigation a subset of five markers were used (sacrum SACR; bilateral thigh xTH2, and tibia xTB2, UWA naming convention), selected for their proximity to the sensor locations in the test-set (Figure~\ref{ch5_figC}), and which made the selection of samples different to earlier studies \citep{5RN558johnson2018tbme,5RN747johnson2018arXiv,5RN490johnson2018mbec}.\par
The test-set was derived from multi data capture sessions conducted between November 2017 to February 2018 at LJMU using Visual3D v6.01 (C-Motion Inc, Germantown, MD). Five Noraxon DTS-3D 518 accelerometers (Noraxon Inc, Scottsdale, AZ) were attached to each of five team-sport athletes (male $80.0~\%$, female $20.0~\%$, height $1.829~\pm~0.080~m$, and mass $75.6~\pm~11.1~kg$) at locations selected for their relevance to an independent study on body segment accelerations (pelvis Pelv; bilateral thigh x\_Th, and shank x\_Sh, LJMU naming convention) \citep{5RN785verheul} (Figure~\ref{ch5_figC}).\par
\begin{figure*}
\begin{adjustwidth}{-2in}{0in}
    \begin{center}
    \includegraphics[width=1.0\linewidth]{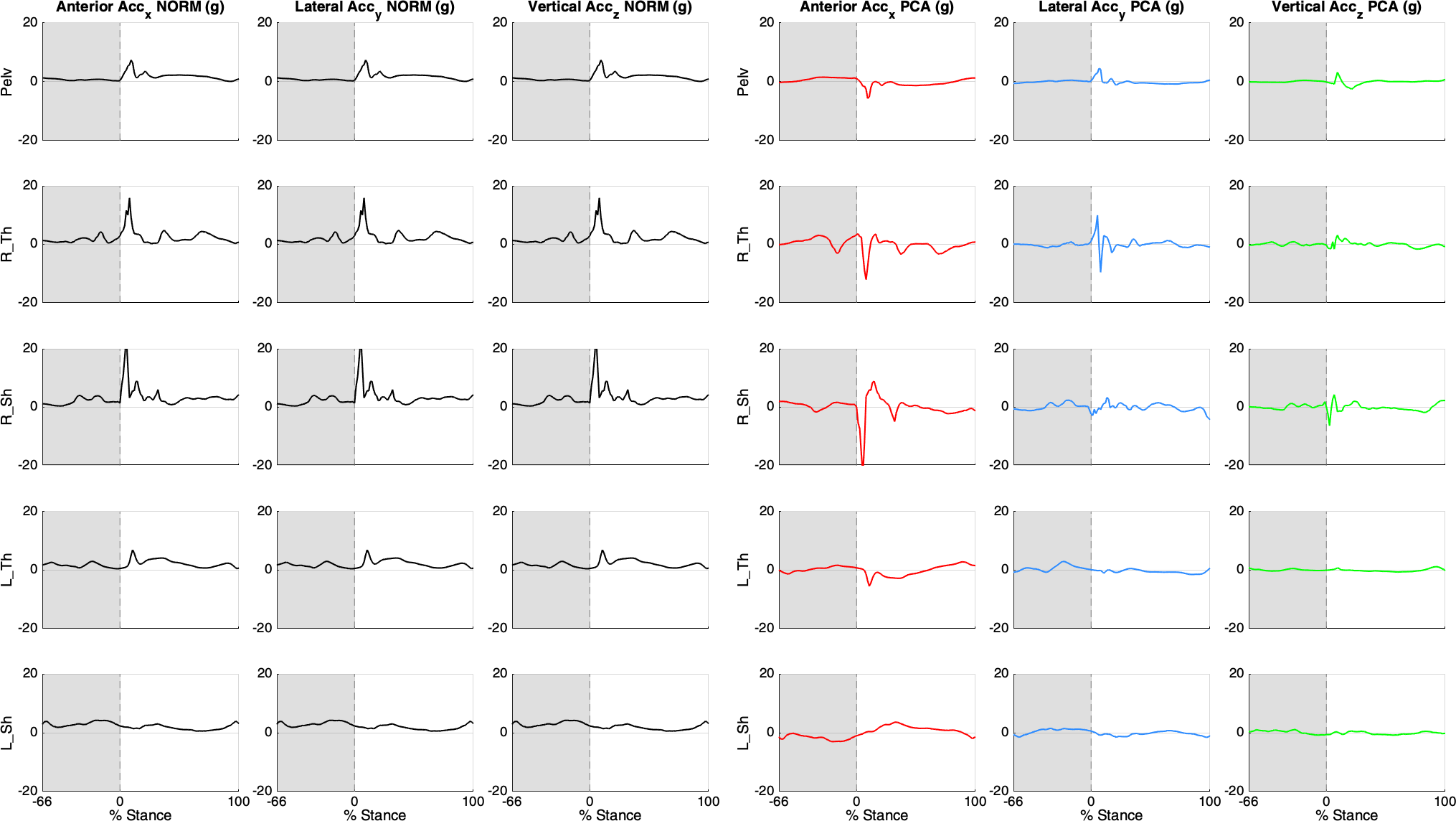}
    \end{center}
    \captionsetup{labelformat=empty}
    \caption{}
    \label{ch5_figD}
\end{adjustwidth}
\end{figure*}

\marginpar{
    \begin{flushleft}
    \vspace{-6.9cm}
    {\color{caption_main}\textbf{Figure \ref{ch5_figD}. Visualization of NORM- (left), and PCA-aligned 3D accelerations (right), sample sidestep right stance limb.}~{\color{caption_sub}Greater signal energy is evident in the stance limb sensors R\_Th and R\_Sh. NORM-aligned accelerations sacrifice dimensionality information and hence the three vectors are identical. PCA-aligned accelerations demonstrate a sweep of information towards Anterior Acc$_x$ (forward, red).}}
    \end{flushleft}
}
\subsection{Data preparation}
Use of the existing data archive was permitted under UWA ethics exemption RA/4/1/8415 (training), and the new data capture was carried out under LJMU ethics approval 17/SPS/043 (test). Data processing was conducted with MATLAB R2017b (MathWorks, Natick, MA) and Python 2.7 (Python Software Foundation, Beaverton, OR), both selected for availability of function libraries.  In the case of MATLAB, for access to the Biomechanical ToolKit 0.3 (Barre and Armand, 2014), and for Python, to conduct low-level image processing using the OpenCV environment, and native HDF5 file handling (\href{https://opencv.org/}{\texttt{\color{blue}{opencv.org}}}, \href{https://www.hdfgroup.org/HDF5/}{\texttt{\color{blue}{hdfgroup.org}}}). The operating system was Ubuntu v16.04 (Canonical, London, UK), running on a desktop PC, Core i7 4GHz CPU, with 32GB RAM and NVIDIA multi-GPU configuration (TITAN X \& TITAN Xp; NVIDIA Corporation, Santa Clara, CA).\par
The data preparation phase was designed to maximize the integrity of the source marker trajectories, sensor accelerations, and force plate data ahead of model training and prediction. The intention was to minimize capture errors (original and new), duplicate files, and select high-quality data rows with labeled marker trajectories (training), sensor accelerations (test), and associated GRF/M. Each trial was normalized to stance phase, and trimmed according to custom lead-in periods to best inform the model as defined by earlier prototypes \citep{5RN587merriaux,5RN593psycharakis,5RN747johnson2018arXiv,5RN490johnson2018mbec}.\par
Basic kinematic templates (based on movement at the sacrum) were used to identify running and sidestepping/cutting in the training and test data (running~$>=$~2.16~$m/s$ \citep{5RN707segers2007}). The sidestepping movement type in particular was selected for its relevance to sporting movements, and knee injury risk, but also for its greater complexity compared with the literature. The majority of trials exhibited right stance limb, with the movement towards the left (a small proportion of sidestepping with crossover technique were removed). The running movement in the test data capture was also sub-categorized into slow (2--3~$m/s$), moderate (4--5~$m/s$), and fast ($>$~6~$m/s$) trials.\par
Registration of a successful foot-strike (FS) onto the force plate, and subsequent toe-off (TO), were both automatically detected using accepted vertical force and stance limb parameters \citep{5RN487milner,5RN589oconnor,5RN602tirosh}, which were then translated to the test accelerations by virtue of synchronized force plate and accelerometer telemetry. The lack of a foot-mounted sensor meant the determination of FS from minimum vertical acceleration at this location was unavailable \citep{5RN507botzel}, and accelerations from the shank sensor were found to be unreliable for this purpose. Identification of the TO gait event from IMU data was considered out of scope for this study, being the primary research objective of other investigations \citep{5RN723allseits,5RN731falbriard,5RN733bertolli}.\par
Virtual IMUs were placed on the mocap skeletons by converting the training-set of marker trajectories into accelerations via double-differentiation, thereby simulating sensor accelerometer data. Since an accelerometer is a free body with an independent coordinate system \citep{5RN745lipton}, in order to model the relationship between 3D accelerations and GRF/M, the accelerations (both those synthesized and recorded) were required to be aligned. Two mathematical methods for automatically re-orienting the accelerations were tested and reported.  The first, was to combine the three directional components into one acceleration magnitude via Euclidean Norm (Figure~\ref{ch5_figD}, left). The second, employed Principal Component Analysis (PCA) via Singular Value Decomposition (SVD), whereby a custom rotation matrix was assembled with the ability to re-orient 3D accelerations in the direction of the greatest PCA energy (\ie forward, for all movement types being investigated). For the training data, a one-off re-orientation was applied by calculating the PCA rotation matrix according to the 3D acceleration at the sacrum location and applying this to all five virtual accelerations. Only one rotation matrix was necessary for the simulated accelerations because their source marker trajectories were aligned with the laboratory global coordinate system.  For the test-set of recorded sensor accelerations, these were all independent and hence an individual rotation matrix was calculated and applied to each. With this test cohort, the effect of PCA can be seen in the sweep of acceleration energy towards the forward (anteroposterior) direction (Figure~\ref{ch5_figD}, right).\par
\subsection{Data representation \& model training}
Model training and prediction was carried out using the Caffe deep learning framework \citep{5RN579jia2014}. Fine-tuning CNN models allows for new investigations with smaller sample sizes to improve their performance by leverage weighting relationships built on earlier training at scale. In deep learning terms, the number of training samples in this study (minimum 1,176, maximum 5,378) was small, and therefore the problem was a candidate for fine-tuning \citep{5RN453krizhevsky2012}. A derivative of the 2012 IVSLRC (\href{http://image-net.org/}{\texttt{\color{blue}{image-net.org}}}) challenge winner AlexNet called CaffeNet had been selected as the strongest model in a similar investigation, and the double-cascade approach (CaffeNet through GRF/M to KJM) had also demonstrated a significant improvement in correlations of $+~4.2~\%$ \citep{5RN558johnson2018tbme,5RN747johnson2018arXiv}. For comparison, and to test a deeper more general model, this investigation also reports a second CNN, ResNet-50, the 2015 IVSLRC challenge winner \citep{5RN752he2015}.\par
\begin{wrapfigure}{l}{0.5\textwidth}
    \includegraphics[width=1.0\linewidth]{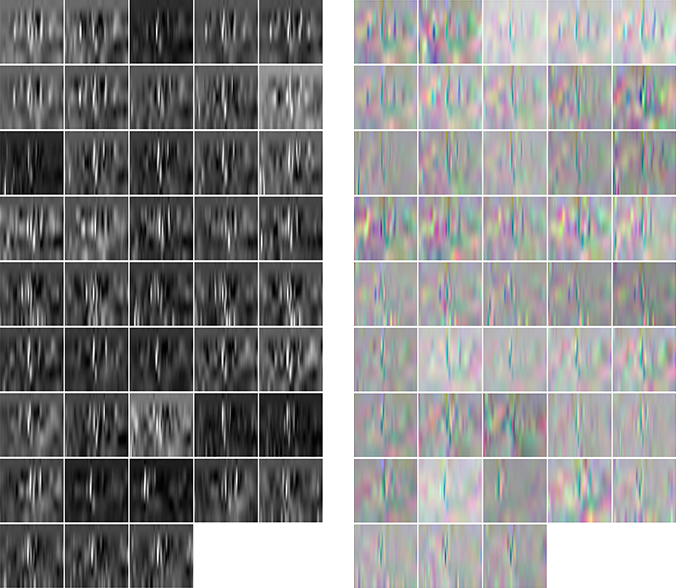}
    \captionsetup{labelformat=empty}
    \caption{}
    \label{ch5_figE}
\end{wrapfigure}

Both AlexNet and ResNet-50 CNN are image classifiers which did not match the required four dimensional input (3D accelerations plus time) and six vector GRF/M waveform output. In order to fine-tune (double-cascade) from these CNN and leverage their existing training, the aligned 4D acceleration inputs were flattened into 2D images by representing the five sensor locations on the horizontal axis, stance-normalized time frames upwards on the vertical axis, and by use of the Python SciPy \href{https://docs.scipy.org/doc/scipy/reference/generated/scipy.misc.imsave.html}{\texttt{\color{blue}{imsave}}} function to map the 3D accelerations onto the RGB colorspace \citep{5RN574du,5RN582ke} (Figure~\ref{ch5_figE}).  Then, so that they would generate GRF/M waveforms (not simply label classifications), the output layer of each CNN was modified from a SoftMax binary to a Euclidean loss layer, which turned the CNN into a multivariate regression network. Most CNNs are classifiers which means the number of features in their output layer is naturally small because it only contains weighting predictions for a discrete set of labels. The high capture frequency of the force plate analog data now being output by the modified network resulted in a non-standard CNN profile (output features $>>$ input features) which was addressed by reducing the number of output features via PCA \citep{5RN558johnson2018tbme}.\par
\marginpar{
    \vspace{0.7cm}
    {\color{caption_main}\textbf{Figure \ref{ch5_figE}. Contact sheets of test accelerations flattened into 2D images.}~{\color{caption_sub}Sidestepping movement combined left and right stance, 43 samples, NORM-aligned accelerations (left) loss of directional information causes monochrome images, PCA-aligned (right) retains color.}}
}
The accuracy and validity of the approach was measured by comparing the correlation of values predicted by the CNN models with the ground truth GRF/M over 100~\% of time-normalized stance. For further comparison, relative root mean squared error $rRMSE$ was reported for individual use-cases \citep{5RN713ren2008}. CNN model predictions were conducted using a single fold of each movement type and stance limb iteration, including an overlaid combination which flipped the left stance limb onto the right, to test the effectiveness of this data augmentation and whether the increase in training samples improved performance. Using the simulated accelerations as the training sets, and the recorded accelerations as the test sets generated variable ratios of training to test samples, however always in favor of the training-set as per convention. For time brevity, single fold experiments were conducted, earlier investigations having demonstrated similarity between single and $k$-fold analysis \citep{5RN558johnson2018tbme}.\par
\sloppy
All CNN models and related digital material supporting this study have been made available  (\href{http://digitalathlete.org}{\texttt{\color{blue}digitalathlete.org}}).\par
\begin{figure*}[t]
\begin{adjustwidth}{-2in}{0in}
    \begin{center}
    \includegraphics[width=1.0\linewidth]{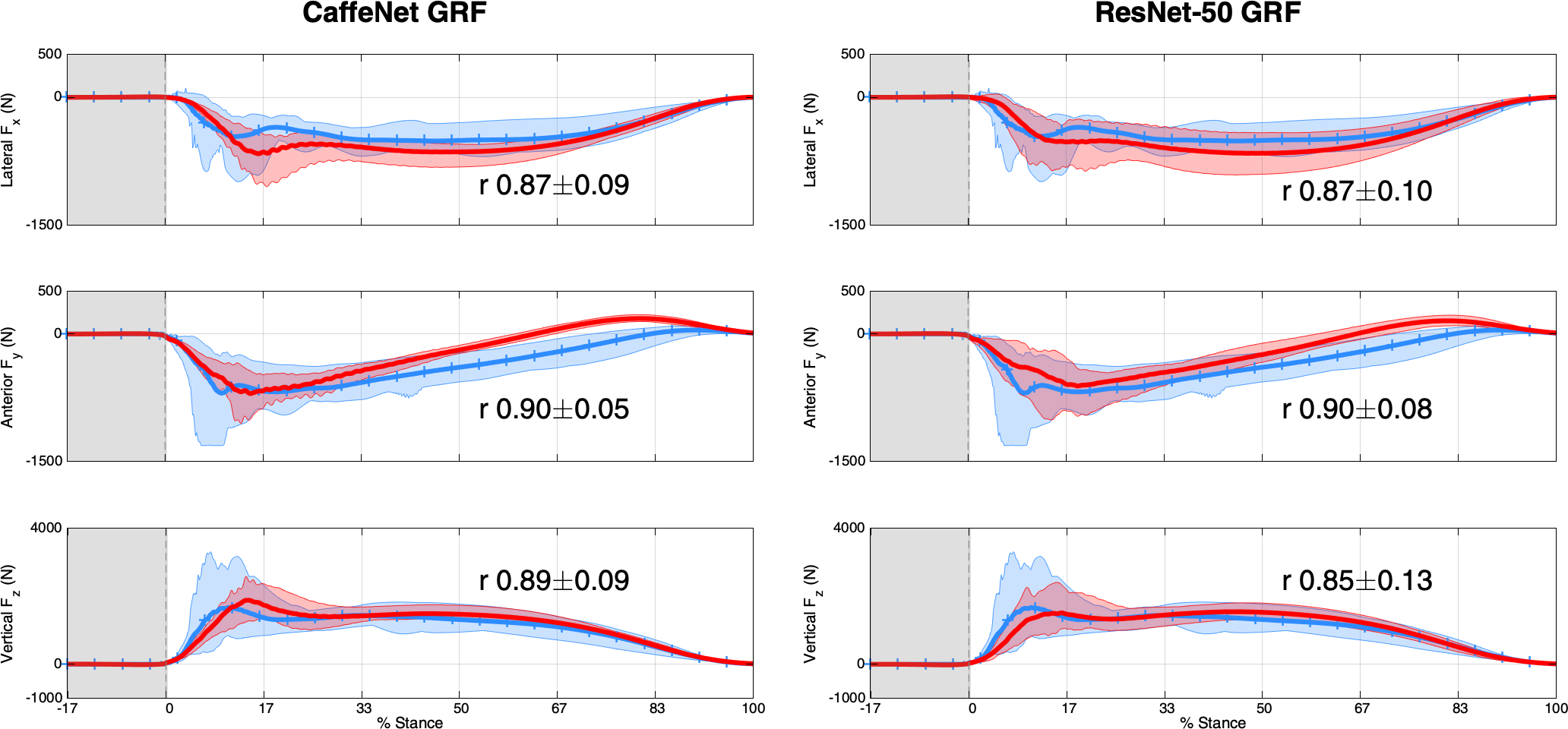}
    \end{center}
    \captionsetup{labelformat=empty}
    \caption{}
    \label{ch5_figF_GRF}
\end{adjustwidth}
\end{figure*}

\fussy
\newpage
\marginpar{
    \begin{flushleft}
    \vspace{0.80cm}
    {\color{caption_main}\textbf{Figure \ref{ch5_figF_GRF}. Ground truth GRF versus predicted response.}~{\color{caption_sub}Test-set ground truth mean GRF (blue, ticks), and predicted response (red), CaffeNet shown left $r(F_{mean})$ 0.89, ResNet-50 right $r(F_{mean})$ 0.87, both double-cascade, interlaced output, correlations over 100~\% stance phase, 25 samples. Cohort selected for strongest $r(F_{mean})$ by CNN (sidestep off the left stance limb), min/max range (shaded areas) and mean (solid lines) depicted.}}
    \end{flushleft}
}
\section{Results}
Compared with ground truth GRF/M, sets of correlations were compared for the two CNN models CaffeNet (Table~\ref{ch5_tabA}) and ResNet-50 (Table~\ref{ch5_tabB}), both modes of acceleration re-orientation Euclidean Norm (accNORM) and alignment by PCA rotation matrix (accPCA), for discrete GRF/M channels $F_x$, $F_y$, $F_z$, $M_x$, $M_y$, $M_z$, and their overall means $F_{mean}$ and $M_{mean}$. Experiments 1.1 and 2.1 list the correlations for the marker to GRF/M models used as seeds for the double-cascade, and are included as reference information.\par
The strongest individual GRF channel correlation was considered first. Across the three GRF channels $F_x$, $F_y$, $F_z$, the highest correlation was found for vertical $F_z$ $0.97~(rRMSE~13.92~\%)$ using CaffeNet (accNORM, experiment 1.8) for moderate speed running off the left stance limb. By channel, anterior $F_y$ was predicted with a correlation up to $0.96~(rRMSE~17.06~\%)$, and lateral $F_x$ $0.87~(rRMSE~21.56~\%)$ both with ResNet-50 off the left stance limb, the former for slow running (accPCA, experiment 2.21), the latter sidestepping (accNORM, experiment 2.24).  Results are shown bolded in their respective tables.\par
\begin{figure*}[t]
\begin{adjustwidth}{-2in}{0in}
    \begin{center}
    \includegraphics[width=1.0\linewidth]{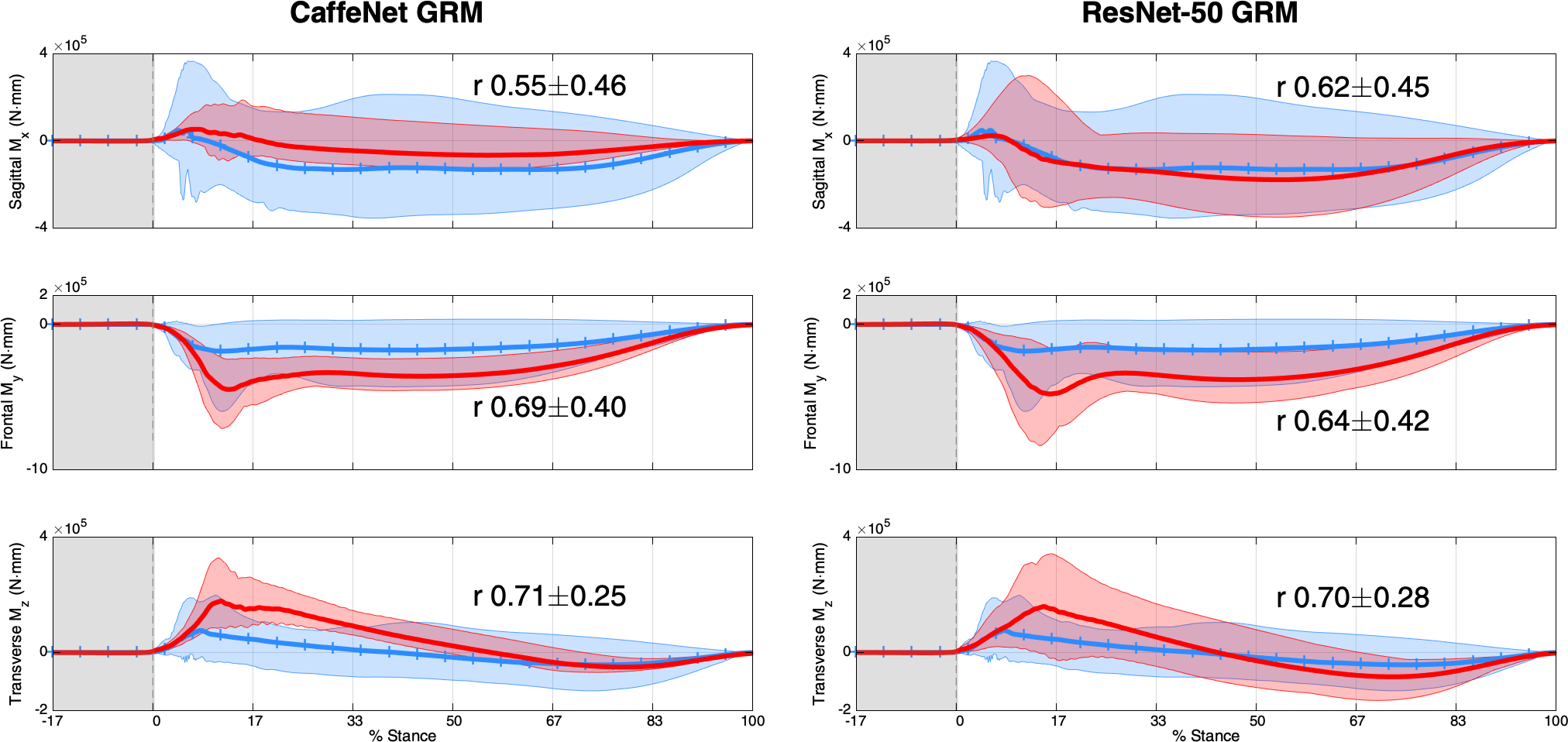}
    \end{center}
    \captionsetup{labelformat=empty}
    \caption{}
    \label{ch5_figF_GRM}
\end{adjustwidth}
\end{figure*}

The mean of the three GRF, $r(F_{mean})$ achieved $0.89$ for CaffeNet (accNORM, experiment 1.24), by comparison, ResNet-50 managed $0.87$ (accNORM, experiment 2.24), both for the same corresponding experiment with a sidestep off the left stance limb (Figures~\ref{ch5_figF_GRF}~\&~\ref{ch5_figG_BA}). The mean of the three GRM, $r(M_{mean})$ proved less than satisfactory, CaffeNet making $0.65$ (accPCA, experiment 1.29), and ResNet-50 $0.65$ (accPCA, experiment 2.29), again both for the same sidestep off the right stance limb (Figures~\ref{ch5_figF_GRM}~\&~\ref{ch5_figG_BA}).\par
\newpage
\marginpar{
    \begin{flushleft}
    \vspace{0.80cm}
    {\color{caption_main}\textbf{Figure \ref{ch5_figF_GRM}. Ground truth GRM versus predicted response.}~{\color{caption_sub}Test-set ground truth mean GRM (blue, ticks), and predicted response (red), CaffeNet shown left $r(M_{mean})$ 0.65, ResNet-50 right $r(M_{mean})$ 0.65, both double-cascade, interlaced output, correlations over 100~\% stance phase, 18 samples. Cohort selected for strongest $r(M_{mean})$ by CNN (sidestep off the right stance limb), min/max range (shaded areas) and mean (solid lines) depicted.}}
    \end{flushleft}
}
\section{Discussion}
Convention dictates that research in the biomechanical sciences is strictly controlled by the primary researcher. The use of broad data sets to train (or fine-tune) deep learning models already breaks this paradigm, but this study went further by inviting a test-set of experiments conducted independently at LJMU, where much of the study design and instrumentation was different to that used for the historical UWA data capture used to train the CNN models.  Performance under these conditions would address the most common criticism that somehow the deep learning model had prior knowledge of test samples (or home-game advantage).\par
As demonstrated by this study, the use of strategies to automatically re-orient 3D accelerations freed the operator from the typical requirements of an initialization posture or sensor calibration. Both the Euclidean Norm and PCA rotation matrix methods solve a major hurdle for adoption in the field while being more elegant than previous solutions \citep{5RN738lebel,5RN741picerno,5RN745lipton,5RN740luinge,5RN727zimmermann,5RN725wouda}. The only drawback being the look-ahead processing requirement which makes either solution \lq{near}\rq real-time, but this is outweighed by the advantages including being agnostic to the direction of participant travel.  With no clear separation of performance characteristics, the two re-orientation methods warrant further investigation, particularly when mathematically the Euclidean Norm solution is more straightforward to implement whereas the PCA approach is a richer source of vector information.\par
\begin{figure*}
\begin{adjustwidth}{-2in}{0in}
    \begin{center}
    \includegraphics[width=0.95\linewidth]{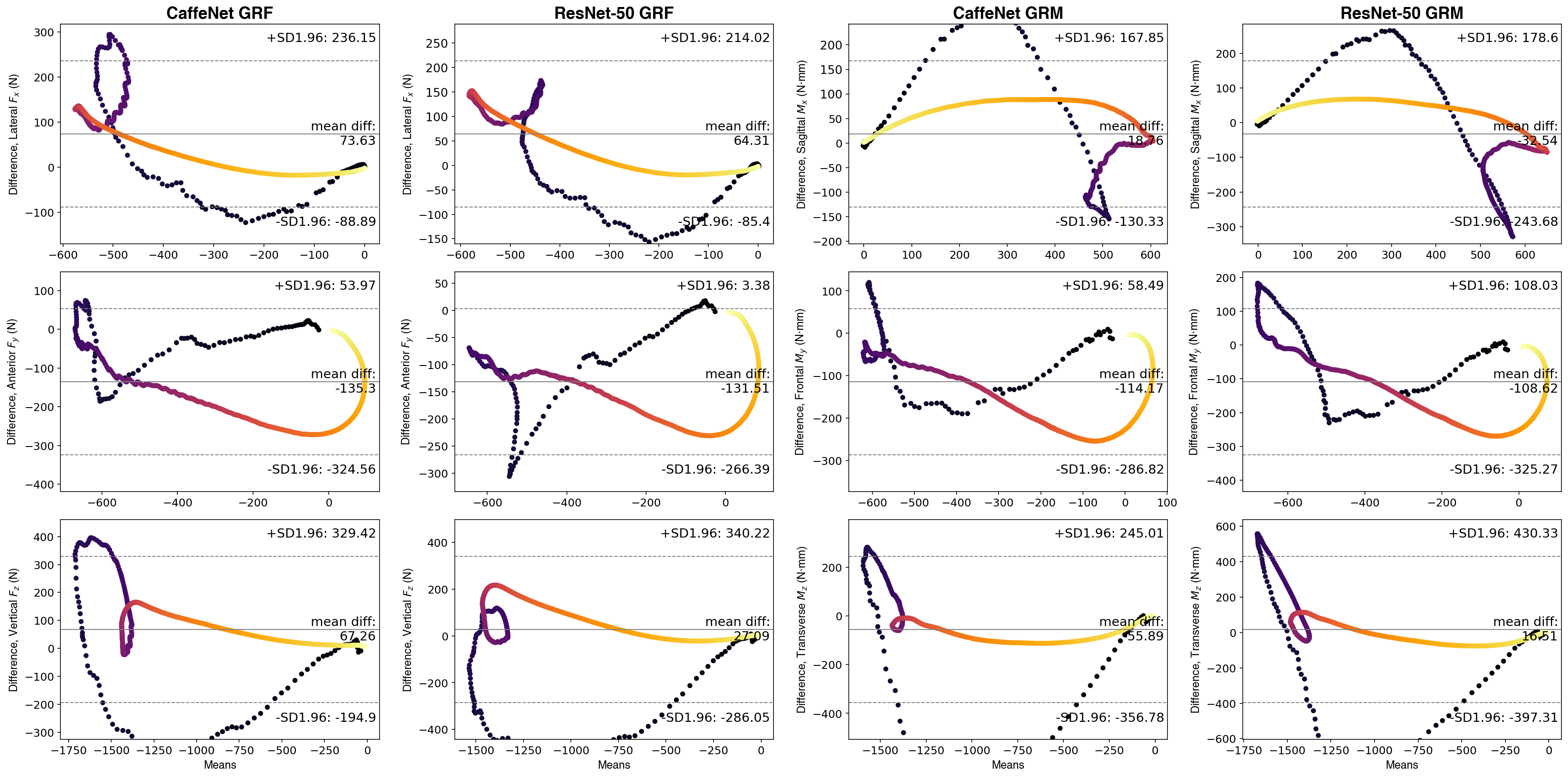}
    \end{center}
    \captionsetup{labelformat=empty}
    \caption{}
    \label{ch5_figG_BA}
\end{adjustwidth}
\end{figure*}

\newpage
\marginpar{
    \begin{flushleft}
    \vspace{0.30cm}
    {\color{caption_main}\textbf{Figure \ref{ch5_figG_BA}. Ground truth GRF/M versus predicted response.}~{\color{caption_sub}Bland-Altman representations of Figures~\ref{ch5_figF_GRF}~\&~\ref{ch5_figF_GRM}. The marker color from dark to light illustrates time from FS.}}
    \end{flushleft}
}
In the competition between the classic CaffeNet model \citep{5RN453krizhevsky2012} and the more recent ResNet-50 \citep{5RN752he2015}, CaffeNet seemed to perform more strongly where there was greater signal strength, \eg $F_z$, $r(F_{mean}$).  ResNet-50, on the other hand, outperformed CaffeNet in conditions of greater noise, \eg $F_x$, which reflects the suitability of the models to each particular use-case, due to either CNN architecture or initial model training. It was theorized that coarse networks like CaffeNet will perform better than deeper networks when the raw source has been blown up to meet the image input requirements, in this case five sensors interpolated to 227 pixels.\par
The LJMU test running data capture was carried out at a number of different speeds and acceleration/deceleration profiles.  In experiments, these were initially grouped by stance limb, and subsequently by a custom L~\&~R combination overlay technique. Time-normalizing the input data according to stance, was expected to reduce the effect of different running speeds, however, variance remained in the results: CaffeNet being the strongest performer, accNORM, running subtypes $r(F_{mean})$ $0.72~\pm~0.10$ (accNORM, experiments 1.2--1.11), $r(F_{mean})$ $0.74~\pm~0.09$ (accPCA, experiments 1.12--1.21); and some of the highest correlations were seen with the samples of running at moderate speed, perhaps due to conformity with the source UWA training data.\par
Mean GRF $F_{mean}$ for ResNet-50 combined stance limb variants outperformed the weakest single limb versions (\eg experiments 2.30 vs 2.22 and 2.23). This is an important finding because a stance-independent model would be far more applicable to game scenarios where the landing limb is unpredictable, and would remove a layer of movement classification hierarchy from the system. The strength of ResNet over CaffeNet in this use-case reflects the preference of deeper CNN architectures to reward greater raw detail with higher learning capacity. This is because these more recent models retain the original size and granularity of the input image through a much longer sequence of convolutions. In other words, ResNet combined L \& R models performed better than a rudimentary mean, and highlights the generalization of the proposed method.\par
\begin{landscape}
\begin{table*}
    \vspace*{-5cm}
    \begin{center}
    \fontsize{6}{6}\selectfont
    \renewcommand{\arraystretch}{1.5}
    \caption[CaffeNet GRF/M component correlations]{\color{caption_main}\textbf{CaffeNet GRF/M component and mean correlations by movement type, stance limb, and motion capture (acceleration orientation) method.}~{\color{caption_sub}CNN double-cascade single fold, output channels interlaced and PCA-reduced, 100~\%~stance.}}
    \label{ch5_tabA}
    \begin{tabular}{|c|c|c|c|c|c|c|c|c|c|c|c|c|c|}
    \hline
    Experiment & Movement & Stance & Motion  & UWA training   & LJMU test     & {$r(F_x)$} & {$r(F_y)$} & {$r(F_z)$} & {$r(M_x)$} & {$r(M_y)$} & {$r(M_z)$} & {$r(F_{mean})$} & {$r(M_{mean})$} \\
    index      & type     & limb   & capture$^2$ & samples        & samples       & & & & & & & &\\
    \hline\hline
    1.1$^1$	&	Sidestep	&	R	&	marker	&	3194	(80.0 \%)	&	798	(20.0 \%)	&	0.98	&	0.98	&	0.99	&	0.97	&	0.97	&	0.98	&	0.98	&	0.98	\\ 
    \hline																														
    1.2	&	Run (acceleration)	&	L	&	accNORM	&	1176	(97.8 \%)	&	27	(2.2 \%)	&	0.25	&	0.57	&	0.93	&	-0.10	&	-0.66	&	-0.21	&	0.58	&	-0.32	\\ 
    1.3	&	Run (acceleration)	&	R	&	accNORM	&	2704	(99.2 \%)	&	22	(0.8 \%)	&	0.27	&	0.59	&	0.84	&	0.00	&	0.70	&	0.24	&	0.57	&	0.32	\\ 
    1.4	&	Run (deceleration)	&	L	&	accNORM	&	1176	(98.6 \%)	&	17	(1.4 \%)	&	0.39	&	0.89	&	0.74	&	0.45	&	-0.14	&	-0.36	&	0.67	&	-0.02	\\ 
    1.5	&	Run (deceleration)	&	R	&	accNORM	&	2704	(99.4 \%)	&	15	(0.6 \%)	&	0.32	&	0.90	&	0.74	&	0.62	&	0.02	&	0.16	&	0.65	&	0.27	\\ 
    1.6	&	Run (fast)	&	L	&	accNORM	&	1176	(97.8 \%)	&	27	(2.2 \%)	&	0.40	&	0.85	&	0.94	&	0.17	&	-0.03	&	0.07	&	0.73	&	0.07	\\ 
    1.7	&	Run (fast)	&	R	&	accNORM	&	2704	(99.0 \%)	&	26	(1.0 \%)	&	0.42	&	0.83	&	0.94	&	0.12	&	0.36	&	0.15	&	0.73	&	0.21	\\ 
    1.8	&	Run (moderate)	&	L	&	accNORM	&	1176	(97.6 \%)	&	29	(2.4 \%)	&	0.56	&	0.95	& \textbf{	0.97	} &	0.58	&	0.23	&	0.40	&	0.83	&	0.40	\\ 
    1.9	&	Run (moderate)	&	R	&	accNORM	&	2704	(98.9 \%)	&	30	(1.1 \%)	&	0.65	&	0.95	&	0.96	&	0.54	&	0.27	&	0.17	&	0.85	&	0.32	\\ 
    1.10	&	Run (slow)	&	L	&	accNORM	&	1176	(98.1 \%)	&	23	(1.9 \%)	&	0.49	&	0.95	&	0.93	&	0.56	&	0.07	&	0.19	&	0.79	&	0.27	\\ 
    1.11	&	Run (slow)	&	R	&	accNORM	&	2704	(98.8 \%)	&	34	(1.2 \%)	&	0.45	&	0.95	&	0.96	&	0.41	&	-0.10	&	-0.14	&	0.79	&	0.05	\\ 
    \hline																														
    1.12	&	Run (acceleration)	&	L	&	accPCA	&	1176	(97.8 \%)	&	27	(2.2 \%)	&	0.33	&	0.61	&	0.92	&	-0.15	&	-0.68	&	0.09	&	0.62	&	-0.25	\\ 
    1.13	&	Run (acceleration)	&	R	&	accPCA	&	2704	(99.2 \%)	&	22	(0.8 \%)	&	0.33	&	0.70	&	0.88	&	0.06	&	0.59	&	0.51	&	0.64	&	0.39	\\ 
    1.14	&	Run (deceleration)	&	L	&	accPCA	&	1176	(98.6 \%)	&	17	(1.4 \%)	&	0.54	&	0.89	&	0.70	&	0.33	&	-0.16	&	-0.30	&	0.71	&	-0.05	\\ 
    1.15	&	Run (deceleration)	&	R	&	accPCA	&	2704	(99.4 \%)	&	15	(0.6 \%)	&	0.40	&	0.88	&	0.71	&	0.74	&	0.09	&	0.30	&	0.66	&	0.38	\\ 
    1.16	&	Run (fast)	&	L	&	accPCA	&	1176	(97.8 \%)	&	27	(2.2 \%)	&	0.33	&	0.86	&	0.94	&	0.13	&	0.21	&	0.26	&	0.71	&	0.20	\\ 
    1.17	&	Run (fast)	&	R	&	accPCA	&	2704	(99.0 \%)	&	26	(1.0 \%)	&	0.35	&	0.85	&	0.95	&	0.32	&	0.22	&	0.02	&	0.71	&	0.19	\\ 
    1.18	&	Run (moderate)	&	L	&	accPCA	&	1176	(97.6 \%)	&	29	(2.4 \%)	&	0.62	&	0.95	&	0.96	&	0.75	&	-0.01	&	0.61	&	0.85	&	0.45	\\ 
    1.19	&	Run (moderate)	&	R	&	accPCA	&	2704	(98.9 \%)	&	30	(1.1 \%)	&	0.62	&	0.95	&	0.96	&	0.59	&	0.21	&	0.25	&	0.84	&	0.35	\\ 
    1.20	&	Run (slow)	&	L	&	accPCA	&	1176	(98.1 \%)	&	23	(1.9 \%)	&	0.57	&	0.95	&	0.92	&	0.45	&	-0.07	&	0.18	&	0.81	&	0.18	\\ 
    1.21	&	Run (slow)	&	R	&	accPCA	&	2704	(98.8 \%)	&	34	(1.2 \%)	&	0.52	&	0.96	&	0.96	&	0.51	&	0.52	&	0.27	&	0.81	&	0.43	\\ 
    \hline																														
    1.22	&	Run	&	L	&	accNORM	&	1176	(90.5 \%)	&	123	(9.5 \%)	&	0.42	&	0.84	&	0.91	&	0.33	&	-0.09	&	0.02	&	0.72	&	0.09	\\ 
    1.23	&	Run	&	R	&	accNORM	&	2704	(95.5 \%)	&	127	(4.5 \%)	&	0.45	&	0.86	&	0.91	&	0.35	&	0.25	&	0.11	&	0.74	&	0.24	\\ 
    1.24	&	Sidestep	&	L	&	accNORM	&	1386	(98.2 \%)	&	25	(1.8 \%)	&	0.87	&	0.90	&	0.89	&	0.57	&	0.32	&	0.63	& \textbf{	0.89	} &	0.51	\\ 
    1.25	&	Sidestep	&	R	&	accNORM	&	3992	(99.6 \%)	&	18	(0.4 \%)	&	0.79	&	0.91	&	0.85	&	0.48	&	0.70	&	0.68	&	0.85	&	0.62	\\ 
    \hline																														
    1.26	&	Run	&	L	&	accPCA	&	1176	(90.5 \%)	&	123	(9.5 \%)	&	0.48	&	0.85	&	0.90	&	0.30	&	-0.14	&	0.22	&	0.75	&	0.13	\\ 
    1.27	&	Run	&	R	&	accPCA	&	2704	(95.5 \%)	&	127	(4.5 \%)	&	0.47	&	0.88	&	0.91	&	0.40	&	0.34	&	0.23	&	0.75	&	0.32	\\ 
    1.28	&	Sidestep	&	L	&	accPCA	&	1386	(98.2 \%)	&	25	(1.8 \%)	&	0.86	&	0.91	&	0.87	&	0.61	&	0.29	&	0.64	&	0.88	&	0.51	\\ 
    1.29	&	Sidestep	&	R	&	accPCA	&	3992	(99.6 \%)	&	18	(0.4 \%)	&	0.79	&	0.89	&	0.85	&	0.55	&	0.69	&	0.71	&	0.84	& \textbf{	0.65	} \\ 
    \hline																														
    1.30	&	Run	&	combined L \& R	&	accNORM	&	3880	(93.9 \%)	&	250	(6.1 \%)	&	0.37	&	0.85	&	0.91	&	0.32	&	0.28	&	0.07	&	0.71	&	0.22	\\ 
    1.31	&	Sidestep	&	combined L \& R	&	accNORM	&	5378	(99.2 \%)	&	43	(0.8 \%)	&	0.84	&	0.91	&	0.88	&	0.59	&	0.49	&	0.03	& \textbf{	0.88	} &	0.37	\\ 
    \hline																														
    1.32	&	Run	&	combined L \& R	&	accPCA	&	3880	(93.9 \%)	&	250	(6.1 \%)	&	0.41	&	0.86	&	0.91	&	0.30	&	0.23	&	0.05	&	0.73	&	0.19	\\ 
    1.33	&	Sidestep	&	combined L \& R	&	accPCA	&	5378	(99.2 \%)	&	43	(0.8 \%)	&	0.82	&	0.91	&	0.83	&	0.69	&	0.44	&	-0.01	&	0.85	&	0.37	\\ 
    \hline
    \end{tabular}
    \end{center}
    {\footnotesize{$^1$Seed for double-cascade model weights.}\\}
    {\footnotesize{$^2$Acceleration re-orientation via Euclidean Norm (accNORM) or PCA rotation matrix (accPCA).}}
\end{table*}    
\end{landscape}
\begin{landscape}
\begin{table*}
    \vspace*{-5cm}
    \begin{center}
    \fontsize{6}{6}\selectfont
    \renewcommand{\arraystretch}{1.5}
    \caption[ResNet-50 GRF/M component correlations]{\color{caption_main}\textbf{ResNet-50 GRF/M component and mean correlations by movement type, stance limb, and motion capture (acceleration orientation) method.}~{\color{caption_sub}CNN double-cascade single fold, output channels interlaced and PCA-reduced, 100~\%~stance.}}
    \label{ch5_tabB}
    \begin{tabular}{|c|c|c|c|c|c|c|c|c|c|c|c|c|c|}
    \hline
    Experiment & Movement & Stance & Motion  & UWA training   & LJMU test     & {$r(F_x)$} & {$r(F_y)$} & {$r(F_z)$} & {$r(M_x)$} & {$r(M_y)$} & {$r(M_z)$} & {$r(F_{mean})$} & {$r(M_{mean})$} \\
    index      & type     & limb   & capture$^2$ & samples        & samples       & & & & & & & &\\
    \hline\hline    
    2.1$^1$	&	Sidestep	&	R	&	marker	&	3194	(80.0 \%)	&	798	(20.0 \%)	&	0.98	&	0.98	&	0.99	&	0.97	&	0.97	&	0.98	&	0.99	&	0.97	\\ 
    \hline																														
    2.2	&	Run (acceleration)	&	L	&	accNORM	&	1176	(97.8 \%)	&	27	(2.2 \%)	&	0.16	&	0.45	&	0.89	&	0.04	&	-0.22	&	0.03	&	0.50	&	-0.05	\\ 
    2.3	&	Run (acceleration)	&	R	&	accNORM	&	2704	(99.2 \%)	&	22	(0.8 \%)	&	0.14	&	0.63	&	0.84	&	0.09	&	0.57	&	0.40	&	0.54	&	0.35	\\ 
    2.4	&	Run (deceleration)	&	L	&	accNORM	&	1176	(98.6 \%)	&	17	(1.4 \%)	&	0.14	&	0.88	&	0.67	&	0.40	&	0.10	&	-0.18	&	0.56	&	0.11	\\ 
    2.5	&	Run (deceleration)	&	R	&	accNORM	&	2704	(99.4 \%)	&	15	(0.6 \%)	&	0.25	&	0.88	&	0.73	&	0.53	&	-0.02	&	0.07	&	0.62	&	0.19	\\ 
    2.6	&	Run (fast)	&	L	&	accNORM	&	1176	(97.8 \%)	&	27	(2.2 \%)	&	-0.04	&	0.84	&	0.93	&	0.26	&	-0.10	&	0.07	&	0.58	&	0.08	\\ 
    2.7	&	Run (fast)	&	R	&	accNORM	&	2704	(99.0 \%)	&	26	(1.0 \%)	&	0.35	&	0.84	&	0.94	&	0.09	&	0.28	&	0.06	&	0.71	&	0.14	\\ 
    2.8	&	Run (moderate)	&	L	&	accNORM	&	1176	(97.6 \%)	&	29	(2.4 \%)	&	0.50	&	0.94	&	0.95	&	0.22	&	-0.02	&	0.15	&	0.80	&	0.12	\\ 
    2.9	&	Run (moderate)	&	R	&	accNORM	&	2704	(98.9 \%)	&	30	(1.1 \%)	&	0.59	&	0.93	&	0.94	&	0.17	&	-0.03	&	-0.11	&	0.82	&	0.01	\\ 
    2.10	&	Run (slow)	&	L	&	accNORM	&	1176	(98.1 \%)	&	23	(1.9 \%)	&	0.48	&	0.94	&	0.89	&	0.66	&	0.03	&	0.17	&	0.77	&	0.29	\\ 
    2.11	&	Run (slow)	&	R	&	accNORM	&	2704	(98.8 \%)	&	34	(1.2 \%)	&	0.51	&	0.95	&	0.94	&	0.55	&	-0.02	&	-0.16	&	0.80	&	0.12	\\ 
    \hline																														
    2.12	&	Run (acceleration)	&	L	&	accPCA	&	1176	(97.8 \%)	&	27	(2.2 \%)	&	0.27	&	0.46	&	0.91	&	-0.13	&	-0.10	&	0.06	&	0.54	&	-0.06	\\ 
    2.13	&	Run (acceleration)	&	R	&	accPCA	&	2704	(99.2 \%)	&	22	(0.8 \%)	&	0.11	&	0.62	&	0.87	&	0.18	&	0.49	&	0.39	&	0.54	&	0.35	\\ 
    2.14	&	Run (deceleration)	&	L	&	accPCA	&	1176	(98.6 \%)	&	17	(1.4 \%)	&	0.28	&	0.89	&	0.74	&	0.40	&	-0.39	&	-0.23	&	0.64	&	-0.08	\\ 
    2.15	&	Run (deceleration)	&	R	&	accPCA	&	2704	(99.4 \%)	&	15	(0.6 \%)	&	0.16	&	0.88	&	0.75	&	0.39	&	-0.05	&	0.01	&	0.60	&	0.11	\\ 
    2.16	&	Run (fast)	&	L	&	accPCA	&	1176	(97.8 \%)	&	27	(2.2 \%)	&	0.02	&	0.80	&	0.92	&	0.07	&	-0.01	&	0.06	&	0.58	&	0.04	\\ 
    2.17	&	Run (fast)	&	R	&	accPCA	&	2704	(99.0 \%)	&	26	(1.0 \%)	&	0.39	&	0.83	&	0.94	&	0.44	&	0.10	&	0.13	&	0.72	&	0.23	\\ 
    2.18	&	Run (moderate)	&	L	&	accPCA	&	1176	(97.6 \%)	&	29	(2.4 \%)	&	0.39	&	0.94	&	0.93	&	0.73	&	0.02	&	0.42	&	0.75	&	0.39	\\ 
    2.19	&	Run (moderate)	&	R	&	accPCA	&	2704	(98.9 \%)	&	30	(1.1 \%)	&	0.53	&	0.95	&	0.96	&	0.47	&	0.29	&	0.15	&	0.81	&	0.30	\\ 
    2.20	&	Run (slow)	&	L	&	accPCA	&	1176	(98.1 \%)	&	23	(1.9 \%)	&	0.22	&	0.92	&	0.89	&	0.62	&	-0.09	&	0.14	&	0.68	&	0.22	\\ 
    2.21	&	Run (slow)	&	R	&	accPCA	&	2704	(98.8 \%)	&	34	(1.2 \%)	&	0.43	& \textbf{	0.96	} &	0.95	&	0.46	&	-0.28	&	-0.19	&	0.78	&	-0.01	\\ 
    \hline																														
    2.22	&	Run	&	L	&	accNORM	&	1176	(90.5 \%)	&	123	(9.5 \%)	&	0.23	&	0.81	&	0.88	&	0.27	&	-0.05	&	0.08	&	0.64	&	0.10	\\ 
    2.23	&	Run	&	R	&	accNORM	&	2704	(95.5 \%)	&	127	(4.5 \%)	&	0.39	&	0.86	&	0.91	&	0.26	&	0.13	&	0.08	&	0.72	&	0.16	\\ 
    2.24	&	Sidestep	&	L	&	accNORM	&	1386	(98.2 \%)	&	25	(1.8 \%)	& \textbf{	0.87	} &	0.90	&	0.85	&	0.70	&	0.27	&	0.56	& \textbf{	0.87	} &	0.51	\\ 
    2.25	&	Sidestep	&	R	&	accNORM	&	3992	(99.6 \%)	&	18	(0.4 \%)	&	0.78	&	0.89	&	0.80	&	0.58	&	0.65	&	0.65	&	0.82	&	0.63	\\ 
    \hline																														
    2.26	&	Run	&	L	&	accPCA	&	1176	(90.5 \%)	&	123	(9.5 \%)	&	0.16	&	0.78	&	0.89	&	0.33	&	-0.07	&	0.08	&	0.61	&	0.11	\\ 
    2.27	&	Run	&	R	&	accPCA	&	2704	(95.5 \%)	&	127	(4.5 \%)	&	0.33	&	0.87	&	0.92	&	0.32	&	0.16	&	0.15	&	0.71	&	0.21	\\ 
    2.28	&	Sidestep	&	L	&	accPCA	&	1386	(98.2 \%)	&	25	(1.8 \%)	&	0.84	&	0.89	&	0.83	&	0.58	&	0.24	&	0.52	&	0.86	&	0.44	\\ 
    2.29	&	Sidestep	&	R	&	accPCA	&	3992	(99.6 \%)	&	18	(0.4 \%)	&	0.75	&	0.88	&	0.82	&	0.61	&	0.63	&	0.70	&	0.82	& \textbf{	0.65	} \\ 
    \hline																														
    2.30	&	Run	&	combined L \& R	&	accNORM	&	3880	(93.9 \%)	&	250	(6.1 \%)	&	0.32	&	0.86	&	0.91	&	0.26	&	0.03	&	0.04	&	0.70	&	0.11	\\ 
    2.31	&	Sidestep	&	combined L \& R	&	accNORM	&	5378	(99.2 \%)	&	43	(0.8 \%)	&	0.82	&	0.90	&	0.82	&	0.51	&	0.44	&	0.01	&	0.85	&	0.32	\\ 
    \hline																														
    2.32	&	Run	&	combined L \& R	&	accPCA	&	3880	(93.9 \%)	&	250	(6.1 \%)	&	0.34	&	0.86	&	0.92	&	0.33	&	0.05	&	0.07	&	0.71	&	0.15	\\ 
    2.33	&	Sidestep	&	combined L \& R	&	accPCA	&	5378	(99.2 \%)	&	43	(0.8 \%)	&	0.81	&	0.91	&	0.84	&	0.53	&	0.43	&	-0.14	&	0.85	&	0.27	\\ 
    \hline																																																										
    \end{tabular}
    \end{center}
    {\footnotesize{$^1$Seed for double-cascade model weights.}\\}
    {\footnotesize{$^2$Acceleration re-orientation via Euclidean Norm (accNORM) or PCA rotation matrix (accPCA).}}
\end{table*}    
\end{landscape}
The major limitation of this study is the selection of sensor locations.  Whereas the shank sensor accelerations were able to successfully identify stance limb (Figure~\ref{ch5_figD}), the vertical acceleration profile at the shank was found to be insufficient to identify the FS event. The lack of mediolateral acceleration energy for running trials was cited for the low $F_x$ and associated mean GRF correlations, due to the CNN model being unable to distinguish signal from noise for these movements. The same symptom of the model misinterpreting noise was considered a contributor to lower GRM correlations. This finding demonstrated the importance of sensors being located as distal as possible in each plane from the center of mass, in order to maximize acceleration profiles, moreover the improvement in correlation performance for sidestepping illustrated the ability of CNN models to distinguish sensor locations by establishing unique internal 3D acceleration signatures. This location awareness is despite a combined acceleration lag and smoothing effect most notable in the response from FS \citep{5RN754pataky2018}, contributed to by the evolution of the workbench code-base from marker-based motion capture input, which down-sampled input accelerations to $250~Hz$, and the proprietary on-board telemetry filtering. The models showed agreement between the two methods ($p~<~0.05$) apart from the difficulty with predicting $F_x$ and $M_x$, $F_z$ and $M_z$, particularly at the FS ramp (Figure~\ref{ch5_figG_BA}, \citep{5RN997bland1986}).\par
Overall, the performance of the deep learning workbench for GRF correlations was impressive when compared with the literature (traditional linear and data science methods) against a hypothesis more demanding than the unidirectional vGRF ($F_z$), movement classification, or counting of steps most commonly investigated \citep{5RN762cust,5RN724jacobs,5RN744hu,5RN736ancillao,5RN737ngoh,5RN727zimmermann,5RN725wouda,5RN722pham,5RN785verheul,5RN732clermont,5RN742thiel,5RN761bertuletti, 5RN999verheul}. 
While the unidirectional two-mass model approach of Clark \etal \citep{5RN998clark2017}, for example, reported encouraging agreement of $R^2=0.95~\pm~0.04$, this was for the vertical force component of running only.  In contrast, the multidimensional mass-spring-damper model investigation by Verheul \etal \citep{5RN999verheul}, reported errors in resultant GRF loading rate of 31~\% during accelerations and 34~\% of RMSE during decelerations. This suggests that the use of physical models has limitations for more dynamic multidimensional team-sport specific tasks, providing further support for the use of a deep learning approach as presented in this paper. 
This study's hypothesis of mean GRF and GRM correlations $>~0.80$ was supported for sidestepping $r(F_{mean})$ regardless of re-orientation methods, CNN models, and stance limb, including the combined experiment 1.31 (CaffeNet, accNORM) which achieved $0.88$. It was noted that the definition of LJMU sidestepping execution at 90\textdegree~was more aggressive than that of UWA at 45--60\textdegree, but that suspected homogeneity in FS pattern inherent to sidestepping with respect to running outweighed any protocol disadvantage.\par
The deep learning workbench employed by this study has demonstrated applicability to biomechanics 4D input and multivariate waveform output. The success of this approach was partly due to the custom nature of the code development, rather than the use of off-the-shelf functions. Plus, these results would not have been possible without headless background batch operation, and on-the-fly generation of CNN architecture and hyperparameter optimization instructions (`prototxt' files) allowing for the drop-in of different models as required.\par
Future investigations should focus on expanding the number of test participants. To improve acceleration signature identification and subsequent model performance, it is strongly recommended to include sensors located at C7 (as typical for team sport) and on each foot. The addition of gyroscope and magnetometer sensor telemetry is expected to increase correlations (the Noraxon sensors used in this study provided 3D linear accelerations only), but would require synthesizing or gathering such information for model training.\par
\section{Conclusions}
A biomechanically relevant system of on-field workload exposure monitoring and acute injury prediction could be a revolutionary contribution to player game preparedness and career longevity. Through a unique ``deep learning workbench for biomechanics'', using legacy marker trajectory trials against new (and independent) accelerometer-driven data capture, the results from this study improve on the literature, but under more challenging sport-related tasks and systematic conditions that make it more relevant for on-field use. Model performance was dependent on gross movement pattern (running or sidestepping) which will be improved by more sophisticated type classification. Both CaffeNet and ResNet-50 demonstrated the ability to profile sensor body location from acceleration signatures. Efforts to address the limitations of no distal sensor location (including C7 and both feet), number of test participants and training samples, and downstream smoothing effects are expected to strengthen the accuracy for all movement types, and particularly the moments of ground kinetics, which could open up this technology for practical application and potentially the prediction of joint kinetics and tissue loading. These results would not have been possible without the multidisciplinary collaboration between sport science and computer science, but the dogma of the invested linear approach and perceived data ownership remain a barrier to adoption. The harvesting of existing team IMU telemetry archives using a deep learning workbench as presented here has the potential to trigger a revolution in the accuracy and validity of wearable sensors from community fitness to professional sport.\par
\section{Acknowledgements}
This project was partially supported by the ARC Discovery Grant DP190102443 and an Australian Government Research Training Program Scholarship. NVIDIA Corporation is gratefully acknowledged for the GPU provision through its Hardware Grant Program, and C-Motion Inc. for the Visual3D licence. Portions of data included in this study were funded by NHMRC grant 400937.\par
\nolinenumbers

\end{document}